\documentclass{article} % For LaTeX2e\
\usepackage{iclr2018_conference,times}
\usepackage[colorlinks=true,citecolor=cyan]{hyperref}%
\usepackage{url}
\usepackage{amsmath,amsfonts,amssymb,amsthm}
\usepackage{bm}
\usepackage{graphicx}
\usepackage{subfigure}
\usepackage{capt-of}
\usepackage{multirow,array}
\usepackage{color}
\usepackage{makecell}
\usepackage{wrapfig}
\usepackage{soul}
\usepackage{authblk}
\usepackage{microtype}
\usepackage{booktabs} 

\makeatletter
\def\maketitle{{%
  \renewenvironment{tabular}[2][]
    {\begin{flushleft}}
    {\end{flushleft}}
  \AB@maketitle}}
\makeatother
\newcommand{\NN}{\mathcal{N}}

\newcommand{\bR}{\mathbb{R}}

\newcommand{\bx}{\bm{x}}

\newcommand{\bth}{\bm{\theta}}

\newcommand{\sgn}[1]{\text{sign}\left(#1\right)}
\newcommand{\ytrue}{y^{\text{true}}}
\newcommand{\ytarget}{y^{\text{target}}}

\DeclareMathOperator*{\argmax}{argmax}

\newcommand{\blue}{\color{blue}}

\usepackage{xcolor}

\title{Understanding and Enhancing the Transferability of Adversarial Examples}

\author[1]{\vspace*{-0.5em}\textbf{Lei Wu}}
\author[2, 3]{\textbf{Zhanxing Zhu}}
\author[2, 3]{\textbf{Cheng Tai}}
\author[2, 3, 4]{\textbf{Weinan E}\vspace*{-0.2em}}

\affil[1]{\footnotesize School of Mathematical Sciences, Peking University, Beijing, China}
\affil[2]{\footnotesize Center for Data Science, Peking University, Beijing, China}
\affil[3]{\footnotesize Beijing Institute of Big Data (BIBDR)}
\affil[4]{\footnotesize Department of Mathematics and PACM, Princeton University, Princeton, NJ, USA}
{
    \makeatletter
    \renewcommand\AB@affilsepx{: \protect\Affilfont}
    \makeatother

    \makeatletter
    \renewcommand\AB@affilsepx{, \protect\Affilfont}
    \makeatother

    \affil[ ]{\texttt{leiwu@pku.edu.cn}}
    \affil[ ]{\texttt{zhanxing.zhu@pku.edu.cn}}
    \affil[ ]{\texttt{chengtai@pku.edu.cn}}
    \affil[ ]{\texttt{weinan@math.princeton.edu}}
}

\iclrfinalcopy % Uncomment for camera-ready version, but NOT for submission.

\begin{document}
\maketitle
\begin{abstract}
	State-of-the-art deep neural networks are known to be vulnerable to adversarial examples,
	formed by applying small but malicious perturbations to the original inputs. Moreover,
	the perturbations can \textit{transfer across models}: adversarial examples
	generated for a specific model will often mislead other unseen models. Consequently
	the adversary can leverage it to attack deployed systems without any query, which severely 
	hinder the application of deep learning, especially in the areas where security is crucial. 

	In this work, we systematically study how two classes of factors that might influence the transferability of adversarial examples.  One is about model-specific factors, including network architecture, model capacity and test accuracy. The other
	is the local smoothness of loss function for constructing adversarial examples. Based on these understanding, a simple but effective strategy is proposed to enhance transferability.  We call it  \emph{variance-reduced attack}, since it utilizes the variance-reduced gradient to generate adversarial example. The effectiveness is confirmed by a variety of experiments on both  CIFAR-10 and ImageNet datasets.
\end{abstract}

%%%%%%%%%%%%%%%%%%%%%%%%%%%%%%%%%%%%%%%%%%%%%%%%
\section{Introduction}
With the resurgence of neural networks, more and more large neural network
models are applied in real-world applications, such as speech recognition, computer vision, etc.
While these models have exhibited  good performance, recent works (\cite{szegedy2013intriguing,goodfellow2014explaining})
show that an adversary is able to fool the model into producing incorrect predictions by manipulating
the inputs maliciously. The corresponding manipulated samples are called \textit{adversarial examples}.
More seriously, it is found that they have cross-model generalization ability,
i.e. the adversarial example generated from one model can fool another different model with a significant probability. We refer to such property as \textit{transferability}. In consequence, hackers can employ this property to attack black-box systems with only limited number of queries (\cite{papernot2016practical,liu2016delving}), inducing serious security issue to deep learning system.

The adversarial vulnerability was first investigated By \cite{szegedy2013intriguing}, in which sophisticated L-BFGS was used to generate adversarial examples. 
Later, a large number of attacks  (\cite{goodfellow2014explaining,papernot2016transferability,liu2016delving,carlini2016towards,chen2017zoo,brendel2017decision}) are proposed. 
Among all the categories of attacks, those based on transferability could be the most dangerous and mysterious, since they surprisingly do not require any input-output query of  the target system. Understanding the mechanism of adversarial transferability could potentially provide various benefits for modern deep learning models. Firstly,  for the already deployed and vulnerable deep neural networks in real systems, it could help to design better strategies to improve the robustness to the transfer-based attacks. Secondly, revealing the mystery behind the transferability of adversarial examples could also extend the existing understandings on modern deep learning, particularly on the effects of model capacity~(\cite{fawzi2015fundamental,madry2017towards}) and model interpretability~(\cite{dong2017towards,ross2017improving}). Therefore, studying the transferability of adversarial example in the context of deep networks is of significant importance.

In this paper, we particularly focus on investigating  two classes of key factors that might influence the adversarial transferability. Inspired by these understandings, we deign a simple but rather effective strategy for enhancing the transferability of adversarial examples.  Our contributions are summarized as follows.

%In this paper, we particularly focus on investigating several key factors that might influence the adversarial transferability, and introduce some new perspectives  on understanding the mechanism of transferability. Inspired by these understandings, we deign a simple but rather effective strategy for enhancing the transferability of adversarial examples, specifically how to choose the best source model and how to construct more transferable adversarial examples.  Our contributions are summarized as follows.
\begin{itemize}
	\item We numerically explore how adversarial transferability relies on the model-specific factors, including the architecture, test accuracy and model capacity. First, it is found that adversarial transfer is not symmetric, i.e. adversarial examples generated from model $A$ can transfer to model $B$ easily does not means the reverse is also natural. This suggests that the explanation based on the similarity of decision boundary is not sufficient, since similarity itself is a symmetric quantity. Second, multi-step attacks seem to outperform one-step attacks in most cases, which is contradictory to the finding in (\cite{kurakin2016adversarial}). Last, we find that adversarial examples generated from a large model appear less transferable than a small model, under the condition they both have good test performance. Interestingly, this finding seems closely related to the previous studies (\cite{madry2017towards,kurakin2016adversarial}) which showed large models are more robust than small ones.

	We also investigate the influence of properties of loss surface. Specifically, our investigation reveals that the local non-smoothness of loss surface harms the transferability of generated adversarial examples. This is consistent with the study of obfuscated gradients (\cite{obfuscated-gradients}) and gradient masking (\cite{tramer2017ensemble}).

	\item Based on previous investigations, we propose a simple but rather effective approach to enhance the transferability.  Inspired by the works (\cite{smilkov2017smoothgrad,balduzzi2017shattered}), we suggest applying the locally averaged gradient instead of the original one to generate adversarial examples. We call it \emph{variance-reduced attack}, since the local averaging have the smoothing effect which suppresses the local oscillation of the loss surface. The effectiveness of our method is justified on both CIFAR-10 and ImageNet datasets, where a large number of state-of-the-art architectures are tested. Different from ensemble-based approaches proposed by \cite{liu2016delving}, our proposal does not require training lots of extra models, which is typically overly expensive in practice. Moreover, we numerically demonstrate that variance-reduced gradient can be combined with ensemble-based and momentum-based (\cite{dongboosting}) approaches seamlessly for producing stronger attacks.  
\end{itemize}

\section{Related Work}

The phenomenon of adversarial transferability was first observed and investigated by \cite{szegedy2013intriguing}. \cite{goodfellow2014explaining} showed that adversarial training can alleviate the transferability slightly, which is recently extended and improved in ( \cite{tramer2017ensemble}) by incorporating to the adversarial examples generated by a large ensemble.
Based on the transferability, \cite{papernot2016practical,papernot2016transferability} proposed a practical black-box attack by training a substitute model with queried information. 
\cite{liu2016delving} showed that targeted transfer is much harder than non-targeted one, and additionally introduced the ensemble-based attacks. More recently, \cite{dongboosting} showed that momentum can help to boost transferability significantly, and by utilizing this property, they won the first-place in \emph{NIPS 2017 Non-targeted Adversarial Attack and Targeted Adversarial Attack} competitions. It is also worth mentioning the work by \cite{moosavi2016universal}, which demonstrated the exists of adversarial perturbations that have cross-sample transferability. Meanwhile, there exist several works trying to explain the adversarial transferability. \cite{papernot2016transferability} contributed it to the similarity between data gradients of source and target models.  By visualization, \cite{liu2016delving} suggested that the transferability comes from the similarity of decision boundaries. \cite{tramer2017space} provided some systematic investigation of this similarity. Unfortunately, similarity is symmetric and it cannot explain our finding that transfer is asymmetric.

Our proposed method enhances the transferability by utilizing informations in the small neighborhood of the clean example, which is inspired by the works on shattered gradient (\cite{balduzzi2017shattered}) and model interpretability (\cite{smilkov2017smoothgrad}). 
Recently, similar strategies are also explored by \cite{obfuscated-gradients} and \cite{he2018decision} for white-box attacks, while we focus on the transferability.
Our method is also related to the work by \cite{athalye2017synthesizing}, which introduced the expectation over transformation (EOT) method to increase robustness. The EOT formulation is similar to our objective~\eqref{eqn: vr-non-target}, which increases the robustness against Gaussian noise. 
\cite{chaudhari2016entropy} suggest that optimizing the landscape smoothed with respect to parameters can lead to solutions generalizing better. Differently, we obtain adversarial perturbation with stronger transferability via smoothing with respect to input $x$.
% Furthermore, our results imply that robustness of adversarial examples and their transferability are strongly correlated. 
% %

%%%%%%%%%%%%%%%%%%%%%%%%%%%%%%%%%%%%%%%%%%%%%%%%
\section{Preliminaries}
\subsection{Adversarial examples}
We use $f(x):  \bR^d \mapsto \bR^K$ to denote the function a model represents, where we omit the dependence on the trainable model parameter $\bth$, since it is fixed in this work.
For many applications of interest, we always have $d \gg 1 $ and $K=O(1)$. According to the local
linear analysis in \cite{goodfellow2014explaining}, it is the high dimensionality that makes
$f(x)$ vulnerable to the adversarial perturbation. That is, considering the $K$-category classification problems as an example, for each $x$, there
exists a  small perturbation $\eta$ that is nearly imperceptible to human eyes, such that the $i$-th output $f_i(x)$ satisfies
\begin{equation}	
	\argmax_{i} f_i(x) = \ytrue, \quad  \argmax_{i} f_i(x+\eta) \neq \ytrue,
	\label{eqn: non-target-def}
\end{equation}
where $\ytrue$ is the true label of the input $x$.  
We call $\eta$ adversarial perturbation and correspondingly $x^{adv}:=x+\eta$ an adversarial
example.

The attack~\eqref{eqn: non-target-def} is called a non-targeted attack since the adversary has no control over  which class  the input $x$ will be misclassified to. 
In contrast, a \emph{targeted attack} aims at fooling the model to produce a wrong label specified by the adversary, i.e.
\[
	\argmax_i f_i(x + \eta) = y^{\text{target}} .
\]

In the \emph{black-box attack} setting, the adversary has no knowledge of the target model (e.g. architecture and parameters) and is not allowed to
query  input-output pairs from the model, i.e. the target model is a pure black-box. However the adversary can
construct adversarial examples on a local model (also called the source model) that is trained
on the same or similar dataset with the target model.  Then it deploys those adversarial examples to fool the target model. %Moreover the success rate of this attack is not negligible due to the transferability of adversarial examples.
This is typically referred to as a black-box attack,
as opposed  to the \emph{white-box} attack whose target  is the source model itself.

\subsection{Models of generating adversarial examples}
In general, crafting adversarial perturbation can be modeled as the following optimization problem,
\begin{equation}
\begin{aligned}
	\text{maximize}_{x'}\quad  & J(x'):= J(f(x'),y^{\text{true}}) \\
		\text{s.t.   }  & \quad \|x'-x\| \leq \varepsilon ,
\end{aligned}
\label{eqn: non-target}
\end{equation}
where $J$ is some loss function measuring the discrepancy between the model prediction and ground truth;
$\|\cdot\|$ is certain norm metric  to quantify the magnitude of the perturbation. For image data, there
is also an implicit constraint: $x' \in [0,255]^d$, with $d$ being the number of pixels.
In practice, the common choice of $J$ is the \textit{cross entropy}.
 \cite{carlini2016towards} introduced a loss function that directly manipulates the output logits instead of probability, which has been also adopted in many works. As to the measurement of distortion, it should be chosen to encourage imperceptibility (\cite{xiao2018spatially}). In this paper,
 the widely used $\ell_{\infty}$ norm is considered.

\paragraph*{Ensemble-based attacks}
To improve the strength of adversarial transferability, instead of using a single model, \cite{liu2016delving} suggested
 using a large ensemble consisting of $f_1,f_2,\cdots,f_Q$ as our source model. Specifically, the non-targeted ensemble-based attack is given by
 \begin{equation}
 	\begin{aligned}
	\text{maximize}_{x'}\quad  & J(\sum_{i=1}^Q w_i f_i(x'),y^{\text{true}}) \\
		\text{s.t.   }  & \quad \|x'-x\| \leq \varepsilon,
\end{aligned}
 \end{equation}
where $w_j$ is the ensemble weight satisfying $\sum_{j=1}^Q w_j=1$. Following (\cite{liu2016delving}), we choose $w_j=1/Q$ in this paper. The targeted counterpart can be derived similarly.

\subsection{Optimizer}
There are various optimizers to solve problem \eqref{eqn: non-target}. Instead of using the classic optimizers, such as Adam, SGD, etc. as in (\cite{liu2016delving,carlini2016towards}), we adopt a more efficient Frank-Wolfe optimizer. In each iteration, we solve the linear approximation of objective function at current solution $\bx_t$ in a constrained space,
\begin{equation}
\begin{aligned}
    s_{t} &= \text{argmax}_{\|s_t\|\leq \varepsilon} J(x_t) + \langle \nabla_x J(x_t), s_t\rangle\\
    x_{t+1} &= \text{proj}_D(x_t + \alpha\, s_t)
\end{aligned}
\label{eqn: frank-wolfe}
\end{equation}
where $\mathcal{D}= [0,255]^d \cap \{x' \,|\, \|x'-x\|\leq \varepsilon\}$ and $\alpha$ is the step size. For $\ell_{\infty}$ norm, the Eqn~\eqref{eqn: frank-wolfe} has an explicit solution given by
\begin{equation}
	x_{t+1} = \text{proj}_D \left(x_{t} + \alpha \text{ sign}(\nabla_x J(x_t)) \right).
	\label{eqn: igsm}
\end{equation}

The attack by evolving ~\eqref{eqn: igsm} for $T$ steps is called \emph{iterative gradient sign method} (\textbf{IGSM}). \cite{kurakin2016adversarial} and \cite{madry2017towards} showed this method can generate  strong adversarial examples. Furthermore, the famous \emph{fast gradient sign method} (\textbf{FGSM}) is a special case with $\alpha=\varepsilon, T=1$. \cite{kurakin2016adversarial} and \cite{tramer2017ensemble} suggested that FGSM can generate more transferable adversarial examples than IGSM. Therefore, both of them are considered in our work.

\paragraph*{Momentum-based approaches}
\cite{dongboosting} recently proposed the momentum-based approaches to enhance adversarial transferability. Specifically, the momentum iterative gradient sign method (\textbf{m-IGSM}) is given by
\begin{equation}
\begin{aligned}
g_{t+1} &= \mu\, g_t + {\nabla J(x_t)}/{\|\nabla J(x_t)\|_1}\\
x_{t+1} & = \text{proj}_D \left(x_{t} + \alpha \text{ sign}(g_t)\right),
\end{aligned}
\end{equation}
where $\mu$ is the decay factor of momentum.

\subsection{Evaluation of Adversarial Transferability}
\label{sec: evaluation-method}
\textbf{Datasets}
To evaluate the transferability, three datasets including MNIST, CIFAR-10 and ImageNet are considered. For ImageNet, directly evaluation on the whole ILSVRC2012 validation dataset is too time-consuming. Therefore, in each experiment we randomly select $5,000$ images that can be correctly recognized by all the examined models.

\textbf{Trained Models}
(i) For MNIST, we trained fully connected networks (FNN) of $D$ hidden layers, with the width of each layer being $500$. For instance, the architecture of FNN width $D=2$ is $784-500-500-10$. All the networks are trained to achieve $100\%$ on training set, and test accuracies of them are higher than $98\%$. (ii) For CIFAR-10, we trained five models: \textit{lenet,resnet20, resnet44, resnet56, densenet}, and test accuracies of them are $76.9\%, 92.4\%,93.7\%, 93.8\%$ and $94.2\%$, respectively. (iii) As for ImageNet, the pre-trained models  are used provided by PyTorch, including \textit{resnet18, resnet34, resnet50, resnet101, resnet152,
vgg11\_bn, vgg13\_bn, vgg16\_bn, vgg19\_bn, densenet121, densenet161, densenet169, densenet201, alexnet, squeezenet1\_1}. 
The Top-1 and Top-5 accuracies of them 
can be found on website\footnote{\url{http://pytorch.org/docs/master/torchvision/models.html}}.  
To increase the reliability of experiments, all the models have been tested. 
However, for a specific experiment we only choose some of them to present since the findings are consistent among the tested models.

\textbf{Criteria}
Given a set of adversarial pairs, 
$\{(x^{adv}_1,\ytrue_1),(x^{adv}_2,\ytrue_2),\dots,(x^{adv}_N,\ytrue_N)\}$,
we calculate their \textit{Top-1 success rates} (\%) fooling a given model $f(x)$  by 
\begin{equation}
	 100\times \frac{1}{N} \sum_{i=1}^N 1[\argmax_i f_i(x^{adv}_i)\neq \ytrue_i].
\end{equation}
If $f(x)$ is the model used to generate adversarial examples, then the rate indicates the  
the white-box attack performance. For targeted attacks, each image $x^{adv}$ is associated with a pre-specified label 
$\ytarget\neq \ytrue$. Then we evaluate the performance of the targeted attack by the following \textit{Top-1 success rate} (\%), 
\begin{equation}
	100 \times \frac{1}{N} \sum_{i=1}^N 1[\argmax_i f_i(x^{adv}_i)=\ytarget_i].
\end{equation}
The corresponding Top-5 rates can be computed  in a similar way.

\textbf{Attacks}
Throughout this paper the cross entropy  \footnote{We also tried the  loss described in~\cite{carlini2016towards}
but did not find its superiority to cross entropy. The reason might be that hard constraints, instead of
 soft penalizations, are used in our formulation.} is chosen as our loss function $J$.
We measure the distortion by  $\ell_{\infty}$ norm, both FGSM and IGSM attacks are considered.

\section{How Model-specific Factors Affect Transferability}
Previous study on transferability mostly focused on the influence of attack methods (\cite{liu2016delving,dongboosting,tramer2017space,kurakin2016adversarial}). However
it is not clear how the choice of source model affects the success rate transferring to target models. In this section, three factors of source model, including architecture, test accuracy and model capacity, are investigated.

\subsection{Architecture}
\label{sec: architecture}
Here we explore how the architecture similarity between source and target model contributes to the transferability. This study is crucial since it can provide us guidance to choose the appropriate source models for effective attacks.  To this end, three popular architectures including ResNet, DenseNet and VGGNet are considered, and for each architecture, two networks are selected. Both FGSM and IGSM attacks are performed on ImageNet dataset. Table~\ref{table: architecture} presents the experiment results, and the choice of hyper-parameter is detailed in the caption.

As we can see, in most cases IGSM outperforms FGSM significantly, especially when the source and target models have the similar architecture.  This observation contradicts the finding by \cite{kurakin2016adversarial} that multi-step attacks are somewhat less transferable than single-step attacks.  However for the attacks from VGGNets to all the ResNets and DenseNets, we do observe IGSM generate adversarial examples weaker than FGSM (blue rates in the figure). So, in general, which method is better should be similarity dependent.

Another finding is that the transfers between two models are non-symmetric, and this phenomenon is more obvious for the models with different architectures. For instance, the success rates of IGSM attacks from \textit{densenet121} to \textit{vgg13\_bn} is $58.7\%$, however the rate from $\textit{vgg13\_bn}$ to $\textit{densenet121}$ has only $19.1\%$. Moreover, the success rates  between models with similar architectures is much higher. For example the success rates of IGSM attacks between \textit{vgg13\_bn} and \textit{vgg16\_bn} are higher than $90\%$,  nearly twice the ones of attacks formed by any other architectures. 
% Also it appears that DenseNet has the strongest attack capability and VGGNet has the weakest attack capability {\color{red} (Necessary to mention this since there is no any insights about this observation?)}. 

%----------------------------------------------------
\begin{table}[!hbt]
\renewcommand{\arraystretch}{1.2}
\centering
\caption{
	Top-1 success rates(\%) of FGSM and IGSM attacks between pairs of models. The row and column denote the source and target model, respectively. For each cell, 
the left is the success rate of FGSM ($\varepsilon=15$), while the right is that of IGSM ($T=5,\alpha=5,\varepsilon=15$).
}
\vspace*{0.5em}
\begin{tabular}{c   |  c     c   |  c     c   |  c     c}\hline
     		& resnet18      & resnet101  & vgg13\_bn     & vgg16\_bn   & densenet121   & densenet161 \\ \hline\hline
resnet18    &  -           &  36.9 / 43.4 &  51.8 / 58.0 & 45.1 / 51.7 & 41.1 / 49.2   &  30.0 / 35.8     \\
resnet101   &  48.5 / 57.2 &  -           &  38.9 / 41.6 & 33.1 / 40.0 & 33.2 / 46.9   &  28.7 / 43.2    \\\hline
vgg13\_bn  	&  {\blue 35.5 / 26.8} & {\blue 14.8 / 10.8}  &  -           & 58.8 / 90.7 & {\blue 19.1 / 15.9}   &  {\blue 13.8 / 11.7}     \\
vgg16\_bn  	&  {\blue 35.2 / 26.1} & {\blue 15.6 / 11.1}  &  61.9 / 91.1 & -           & {\blue 21.1 / 16.8}   &  {\blue 15.8 / 13.2}    \\\hline
densenet121 &  49.3 / 63.8 & 34.4 / 50.7  &  47.6 / 58.7 & 41.0 / 57.8 & -             &  38.5 / 73.6    \\
densenet161 &  45.7 / 56.3 & 33.8 / 54.6  &  48.6 / 56.0 & 41.3 / 55.9 & 43.4 / 78.5   &  -      \\ \hline
\end{tabular}
\label{table: architecture}
\end{table}
%----------------------------------------------------

\subsection{Model Capacity and Test Accuracy}
\label{sec: capacity}
\begin{figure}[!ht]
\centering
\includegraphics[width=0.9\textwidth]{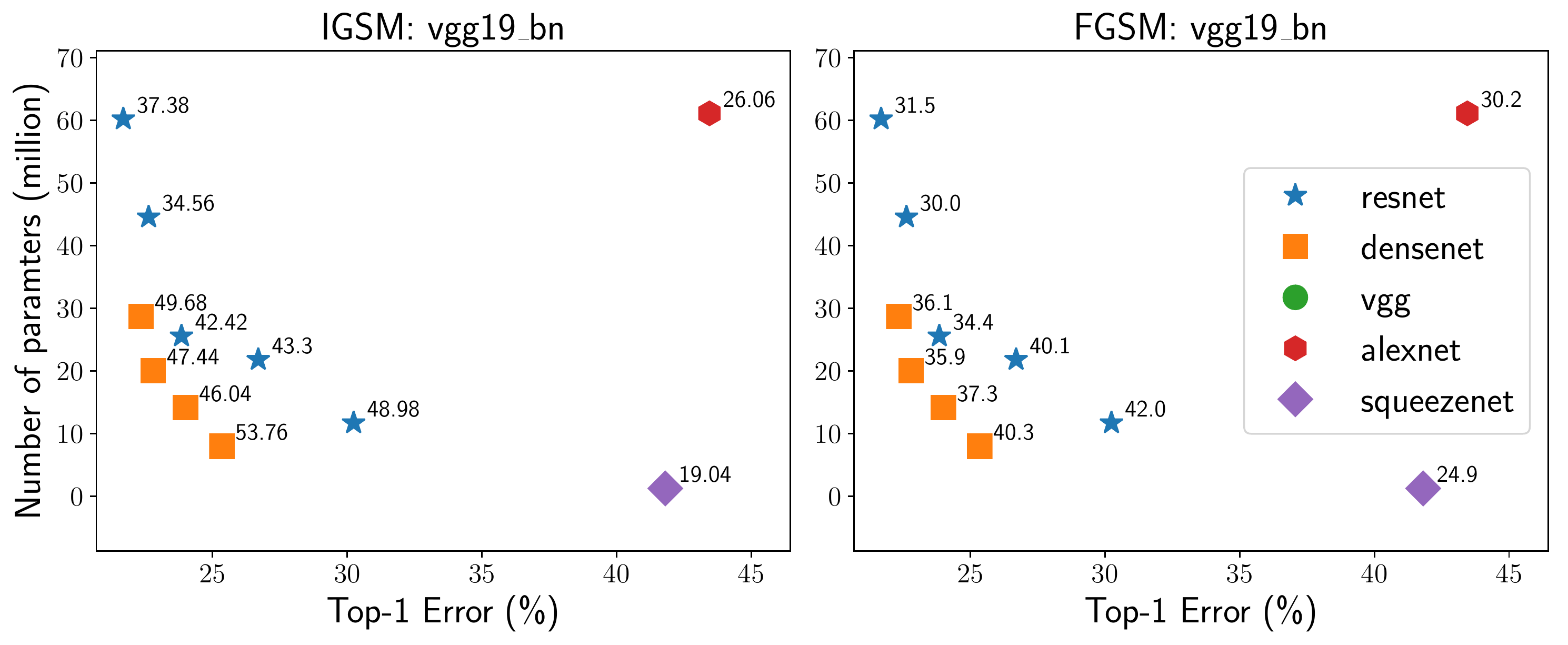}
\caption{Top-1 success rates of FGSM($\varepsilon=15$) and IGSM ($k=20,\alpha=5,\varepsilon=15$) attacks against \textit{vgg19\_bn} for various models. The annotated value
is the percentage of adversarial examples that can transfer to the \textit{vgg\_19}. Here, the models of vgg-style have been removed to exclude the influence of architecture similarity. }
\label{fig: complexity-accuracy-foolingrate}
\end{figure}

We first study this problem in ImageNet dataset, with \textit{vgg19\_bn}\footnote{We also tried other nets, and the results show no difference.} chosen as our target model.  A variety of models are used as source models to perform both FGSM and IGSM attacks. 
The results are displayed in Figure~\ref{fig: complexity-accuracy-foolingrate}.
The horizontal axis is the Top-1 test error, while the vertical axis is the number of model parameters
that roughly quantifies the model capacity.
We can see that   the  models with powerful attack capability
concentrate in the bottom left corner, and the fooling rates are much lower for those models  with either large test error or large number of parameters.

The decision boundaries of high-accuracy models should be similar, since all of them approximate the ground-truth decision boundary of data very well. On the contrary, a low-accuracy model must has a decision boundary relatively different from the high-accuracy models. Therefore it is not surprising to observe that \textbf{high-accuracy models tend to exhibit stronger attack capability}. 

However it is somewhat strange that \textbf{adversarial examples generated from deeper model appear less transferable}.  To further confirm this observation, we conduct additional experiments on MNIST and CIFAR-10. Table~\ref{table: capcity-success_rate} shows the results, which is basically consistent with previous observation, though there exists counter-examples, attacks from \textit{resnet44,resnet56} to \textit{resnet20}. It suggests us not to use deep models as the source models when performing transfer-based attacks, even though we still can not fully understand it. It is also worth to mention that some works (\cite{kurakin2016adversarial,madry2017towards}) observed a dual phenomenon that deeper models are more robust against adversarial perturbations.  

\begin{table}[!htb]
\renewcommand{\arraystretch}{1.2}
\centering
\caption{
	Each cell ($S$,$T$) denotes the Top-1 success rate of attack from  source model $S$ to target model $T$.
}
\subtable[MNIST, FGSM attack with $\varepsilon=40$]{
\begin{tabular}{|c|cccc|}
\hline
	 & $D=0$ & $D=2$ & $D=4$ & $D=8$ \\\hline
 $D=0$ & - & 62.9 & 62.9 & 64.4 \\
 $D=2$ & 52.9 & - & 48.3 & 49.4 \\
 $D=4$ & 47.3 & 43.1 & - & 44.8 \\
 $D=8$ & 31.2 & 29.2 &  29.0 & - \\\hline
\end{tabular}
}
\subtable[CIFAR-10, FGSM attack with $\varepsilon=10$]{
	\begin{tabular}{|c|cccc|}
	\hline
			& resnet20 & resnet44 & resnet56 & densenet \\\hline
	resnet20 & -    & 70.4 & 64.0 & 71.6\\
	resnet44 & 65.4 & -  & 57.1 & 65.8\\
	resnet56 & 66.2 & 62.9 &-  & 40.3\\
	\hline
	\end{tabular}
}
\label{table: capcity-success_rate}
\end{table}

\section{Shattered Gradients}
In previous section, we explore how model-specific factors influence the transferability of adversarial examples. In this part, we systematically investigate how the smoothness of $J(x)$ impacts the transferability across models. 

\begin{figure}[!htb]
\centering
\subfigure[]{
\raisebox{0.0mm}{\includegraphics[width=0.22\textwidth]{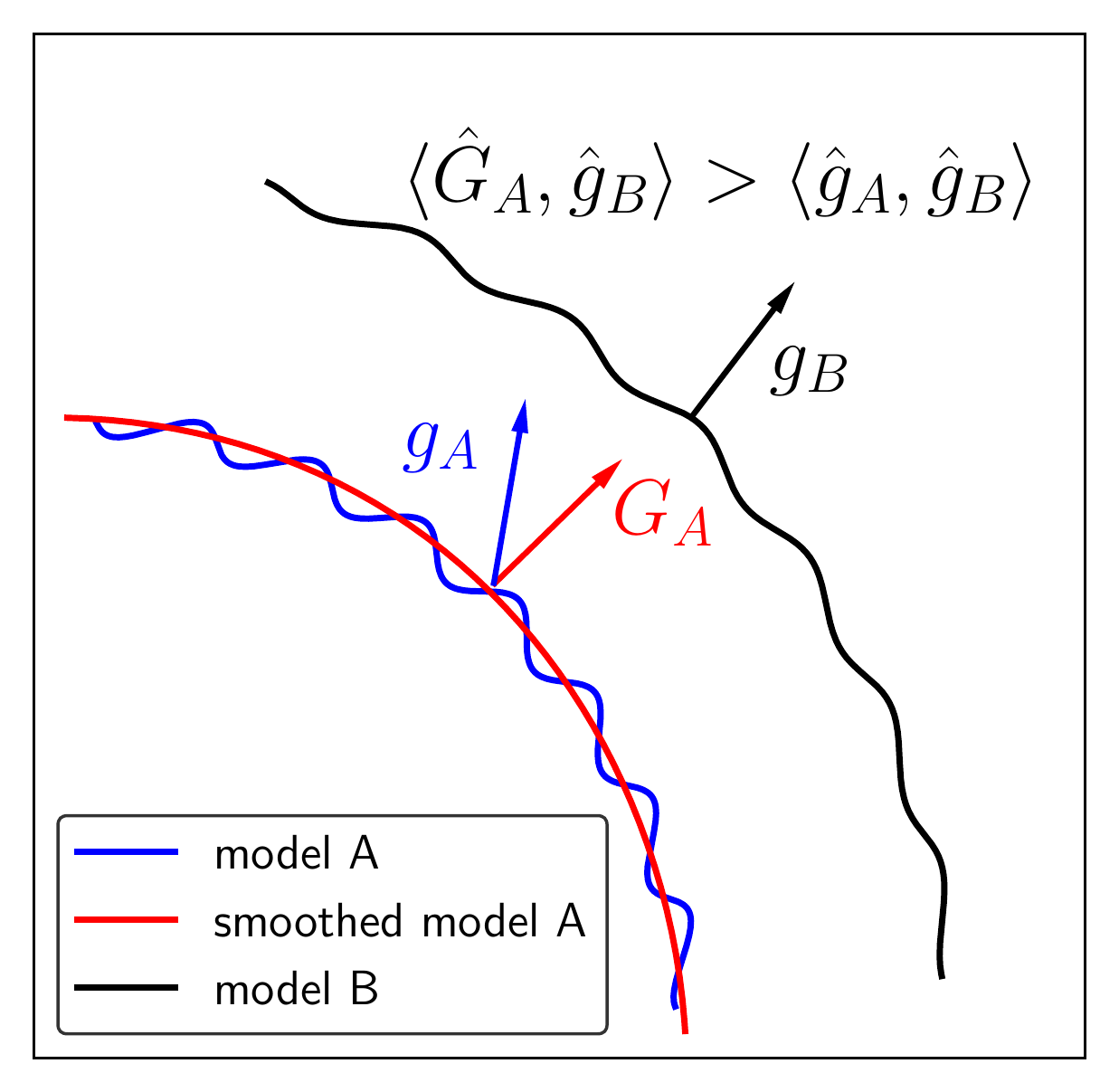}}
\label{fig: cartoon}
}
\subfigure[]{
\raisebox{-3.0mm}{\includegraphics[width=0.25\textwidth]{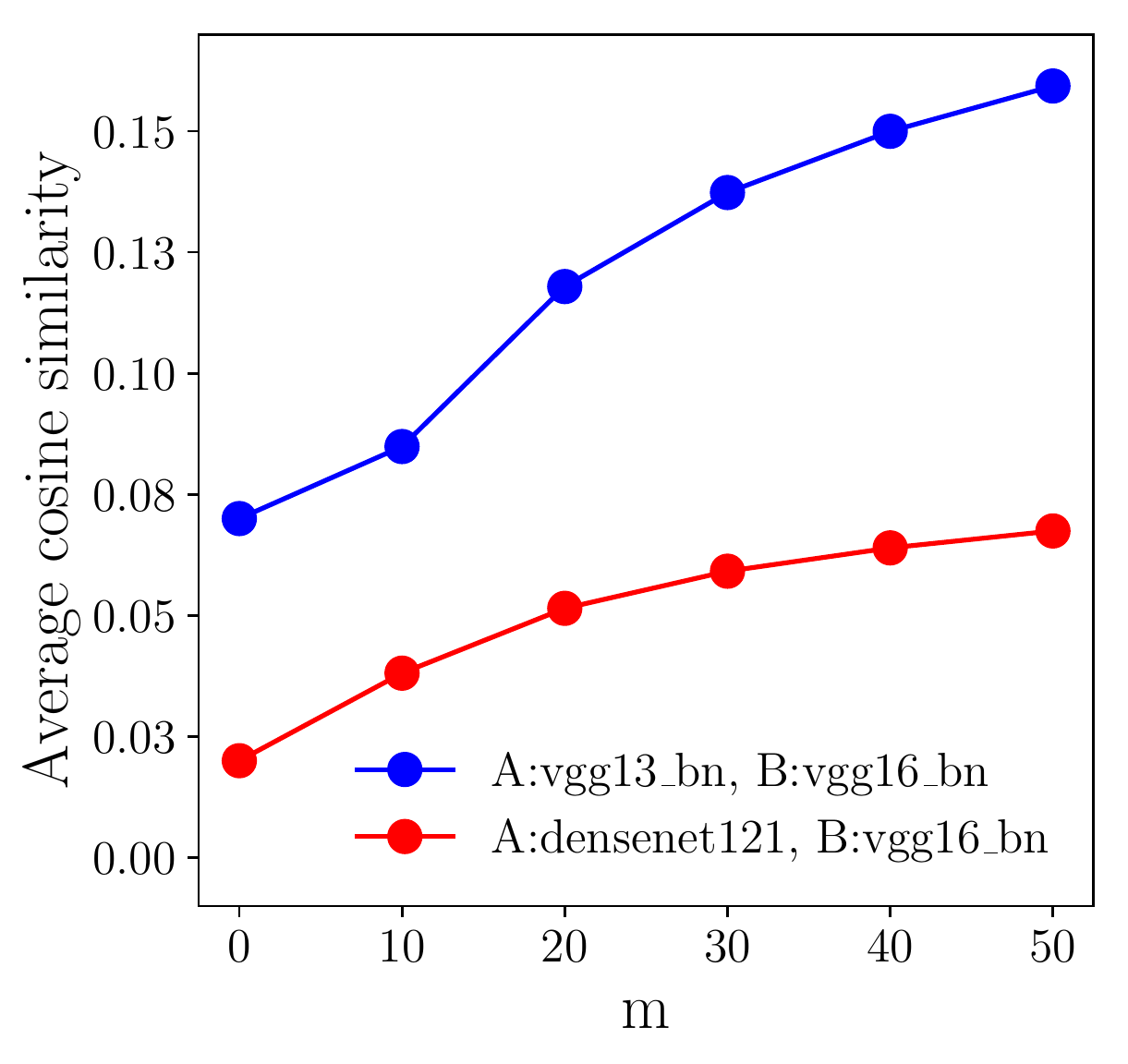}}
\label{fig: cosine_smilarity}
}
\subfigure[]{
	\includegraphics[width=0.43\textwidth]{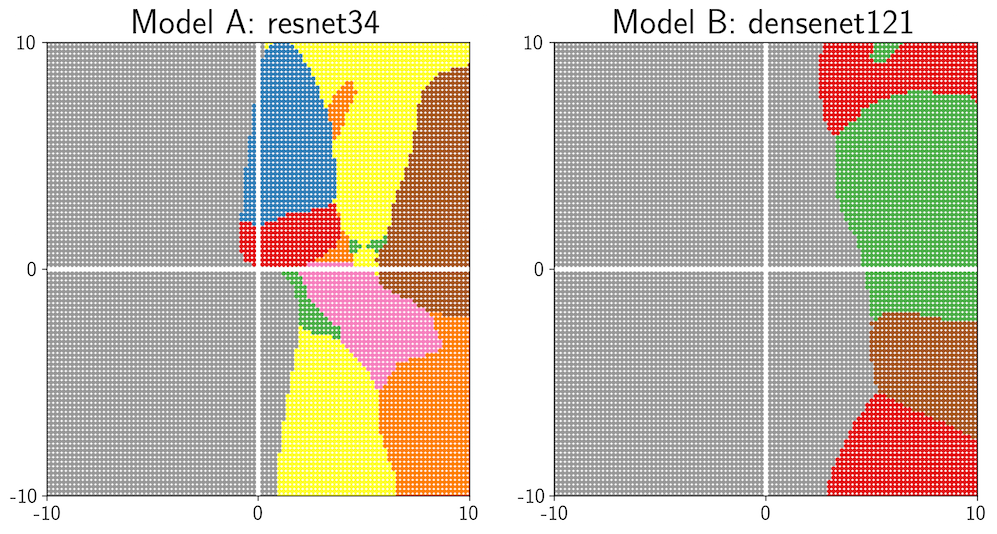}
\label{fig: decision-boundary}
}
\caption{(a) The three solid lines denote the level set of landscapes. The hatted vector denotes the corresponding unit vector. (b) Cosine similarity between the gradients of source and target models. (c) Visualization of decision boundaries. The origin corresponds to the clean image shown in Figure~\ref{fig: sample-image} of Appendix.}
% \label{fig: cartoon}
\end{figure}

For simplicity, assume model A and B are the source and target models, respectively and let $g(x):=\nabla J(x)$ be the gradient. Previous methods use $g_A(x)$  to generate adversarial perturbations, so the transferability mainly depends on how much instability of $g_A$ can transfer to model B. The shattered gradients phenomenon studied in (\cite{balduzzi2017shattered}) implies that gradient $g_A$ is very noisy, though the model is trained on training set very well. We hypothesize that the noise hurts the transferability of $g_A$, as illustrated in cartoon Figure~\ref{fig: cartoon}. Since both model A and B have very high test accuracy,  their level sets should be similar globally, and $J_B$ is probably unstable along $g_A$. However, as illustrated in the figure, the local fluctuation of $g_A$ make the sensitivity less transferable. One way to alleviate it is to smooth the landscape $J_A$, thereby yielding a more transferable direction $G_A$, i.e. $\langle \hat{G}_A,\hat{g}_B\rangle > \langle \hat{g}_A,\hat{g}_B\rangle$.

Inspired by  the standard technique from distribution theory, we suggest to  smooth the landscape as follows,
\begin{equation}
J_{\sigma}(x):= \int J(x-x') \psi_{\sigma}(x') dx',
\end{equation}
where $J_{\sigma}$ is the smoothed landscape, and the smoothing is achieved by convolution with a mollifier, defined as a smooth function satisfying,
\[
\psi_{\sigma}(x)=\sigma^{-d}\psi(x/\sigma) \qquad \int_{\mathbb{R}^d} \psi(x) dx =1.
\]
In this paper, the Gaussian mollifier, i.e. $\psi(x) = \frac{1}{(2\pi)^{d}} e^{-\|x\|^2/2}$, is used, and the  corresponding gradient can be calculated by 
\begin{equation}
	G_{\sigma}(x) = \mathbb{E}_{\xi\sim\mathcal{N}(0,\sigma^2)} [g(x+\xi)].
	\label{eqn: smooth-grad}
\end{equation}
This strategy has been employed by \cite{smilkov2017smoothgrad} to improve the interpretability of gradient saliency maps, which demonstrates that the local averaged gradient $G_{\sigma}(x)$ is more informative and interpretable than $g(x)$.  Here, we further show that \emph{$G_{\sigma}(x)$ is more transferable than $g(x)$} by numerical experiments on ImageNet. 

We first quantify the cosine similarity between gradients of source and target models, respectively. Two attacks are considered:  \textit{vgg13\_bn}$\rightarrow$\textit{vgg16\_bn}, \textit{densenet121}$\rightarrow$\textit{vgg13\_bn}, which represent the within-architecture and cross-architecture transfer, respectively. We choose $\sigma=15$, and the expectation in \eqref{eqn: smooth-grad} is estimated by using $\frac{1}{m}\sum_{i=1}^m g(x+\xi_i)$.  To verify the averaged gradients do transfer better, we plot the cosine similarity between source and target model versus the number of samples $m$. Figure~\ref{fig: cosine_smilarity} display the average similarity calculated from $5,000$ randomly selected samples, which shows that the cosine similarity between $G_A$ and $g_B$ are indeed larger than the one between $g_A$ and $g_B$. As expected, the similarity increases with $m$ monotonically, which further justifies that $G_A$ is more transferable than $g_A$. 

Next, we visualize the transferability in Figure~\ref{fig: decision-boundary} by comparing the decision boundaries of \emph{resnet34} (model A, the source model) and \emph{densenet121} (model B, the target model). The horizontal axis represents the direction of $G_A$ of \emph{resnet34},  estimated by $m=1000,\sigma=15$, and the vertical axis denotes orthogonal direction $h_A:= g_A - \langle g_A,\hat{G}_A\rangle \hat{G}_A$.  Each point in the 2-D plane corresponds to the image perturbed by $u$ and $v$ along each direction,
\[
	\text{clip}(x + u \, \hat{G}_A + v\, \hat{h}_A,0,255),
\]	
where $x$ is the clean image. 
 It can be easily observed that for \emph{resnet34} (model A), a small perturbation
in both directions can produce wrong classification. However, when applied to \emph{densenet121} (model B), a large perturbation along  $h_A$ cannot change the classification result, while a small perturbation along $G_A$ can change the prediction easily.  This further confirms that local averaging indeed extracts the more transferable part of $g_A$.

The above analysis implies that the local oscillation of loss surface do harm the transferability of adversarial examples. This is similar to the phenomenon of gradient masking observed in the defense of adversarial example (\cite{tramer2017ensemble,papernot2016towards,obfuscated-gradients}). 
Our experiments suggest that gradient masking  also exists for normally trained models to some extent, i.e. shattered gradients. The noise contained in the shattered gradient makes the generated adversarial examples less transferable.

\section{Variance-reduced Attack}
\subsection{Method}
\label{sec: method}
Previous study suggests to alleviate the shattering of gradients by optimizing the smoothed loss function,
\begin{equation}
\begin{aligned}
	\text{maximize}\quad  & J_{\sigma}(x'):= 
				\mathbb{E}_{\xi\sim\mathcal{N}(0,\sigma^2)} [J(x'+\xi)] \\
		\text{s.t.   }  & \quad \|x'-x\| \leq \varepsilon .
\end{aligned}
\label{eqn: vr-non-target}
\end{equation}
Intuitively, this method can also be interpreted as  generating adversarial examples that are robust to  Gaussian perturbation. We expect that the generated robust adversarial examples can still survive easily in spite of the distinction between source and target model.   

Using the iterative gradient sign method to solve \eqref{eqn: vr-non-target} yields the following iteration formula:
\begin{equation}
\begin{aligned}
	G_t &= \frac{1}{m}\sum_{i=1}^m \nabla J(x^t+\xi_i), \quad \xi_i\sim \NN(0,\sigma^2I) \\
 	x_{t+1} &= \text{proj}_D\left(x_t + \alpha \, \sgn{G_t}\right),
\end{aligned}
\label{eqn: vr-igsm}
\end{equation}
where $G_t$ is a mini-batch approximation of $\mathbb{E}_{\xi\sim\mathcal{N}(0,\sigma^2I)}[\nabla J(x+\xi)]$. Compared to IGSM, the gradient is replaced by an  averaged one, which removes the local fluctuation. Therefore we call this method variance-reduced iterative gradient sign method (vr-IGSM). The special case $T=1,\alpha=\varepsilon$, is accordingly called variance-reduced fast gradient sign method (vr-FGSM).
For any other optimizer, the corresponding variance-reduced version can be derived similarly. However, in this paper we only consider IGSM and FGSM for simplicity.

%---------------------------------------------

\subsection{Effectiveness}
In this section, we evaluate the effectiveness over both CIFAR-10 and ImageNet datasets.  

\begin{table}[!hbt]
\renewcommand{\arraystretch}{1.2}
\centering
\caption{Top-1 success rates (\%) of attacks between pairs of models. The row and column denote the source and target model, respectively.  For each cell,
the left  is the success rate of the normal attack the right is that of
corresponding variance-reduced attack. In this experiment, distortion $\varepsilon = 10$ }
\subtable[FGSM versus vr-FGSM ($\sigma=15,m=100$)]{
\begin{tabular}{|c|c c c|}
\hline
         & lenet & resnet-20 & densenet \\\hline
 lenet   & -  &29.0 / 28.7 & 28.7 / 29.0\\
 resnet-20 &25.4 / 30.3 & -& 71.6 / 90.0\\
 densenet & 26.0 / 31.9&72.0 / 90.5 & -\\\hline
\end{tabular}
}
\subtable[IGSM($T=5,\alpha=4$) versus vr-IGSM($T=5,\alpha=4,m=20,\sigma=15$)]{
\begin{tabular}{|c|c c c|}
\hline
         & lenet & resnet-20 & densenet \\\hline
 lenet   & -  &30.8 / 31.7 & 30.6 / 31.3\\
 resnet-20 &24.9 / 26.9  &-& \textbf{85.9 / 97.6}\\
 densenet & 25.9 / 28.4  &\textbf{92.9 / 99.0} & -\\\hline
\end{tabular}
}

\label{tab: cifar10-fgsm}
\end{table} 

\subsubsection{CIFAR-10}
We first consider the adversarial transferability among three models: \emph{lenet}, \emph{resnet-20} and \emph{densenet}. The test accuracies of them are $76.9\%$, $92.4\%$ and $94.2\%$ respectively. Table~\ref{tab: cifar10-fgsm} shows, in general, variance-reduced gradient indeed improve the transfer rate consistently. In particular, the transferability between \emph{resnet-20} and \emph{densenet} are enhanced significantly. However the variance-reduced gradient does not help too much for the transfers from \emph{lenet} to \emph{resnet-20} and \emph{densenet}.  It is interesting to note that \emph{lenet} has very weak attack capability, which probably is due to its low accuracy. It does not learn the data very well, causing that its decision boundary be very different from the high-accuracy model \emph{resnet20} and \emph{densenet}. This is consistent with our finding in Section~\ref{sec: capacity}.

\subsubsection{ImageNet}
\label{sec: imagenet}
Previous experiments show that our method indeed can improve the success rate. We now turn to the more real dataset, imageNet. 
To make the experimental result reliable, both one-step and multi-step methods, targeted and non-targeted attacks, single-model and ensemble-model based approaches are examined.

\paragraph*{Single-model based approaches}
Here we test both one-step and multi-step attacks, specifically for multi-step attacks, we make the number of gradient calculation per sample be fixed $100$ for a fair comparison. The results of multi-step attacks are shown in Table~\ref{table: non-target-igsm} (readers can refer to Table~\ref{table: non-target-fgsm} in Appendix for the result of one-step attack). As we can see, our method enhance the transferability dramatically for all the attacks. Please especially note those bold rates, where the improvement has reached about $30\%$.

%----------------------------------------------------
\begin{table}[!ht]
\renewcommand{\arraystretch}{1.2}
\centering
\caption{
	Top-1 success rates(\%) of non-targeted IGSM and vr-IGSM attacks between pairs of models. The row and column denote the source and target model, respectively. For each cell, the left is the success rate of IGSM ($T=100,\alpha=1$), while the right is the that of vr-IGSM ($m=20,\sigma=15,T=5,\alpha=5$). In this experiment, distortion $\varepsilon=15$.
}
\vspace*{0.5em}
\begin{tabular}{c   |  c     c     c   |  c     c}\hline
      & densenet121 & resnet152 & resnet34 & vgg13\_bn & vgg19\_bn\\ \hline\hline
densenet121  &  -   &  \textbf{50.1 / 80.6}   & 59.9 / 87.2   &  62.2 / 82.2    &  56.5 /84.3   \\
resnet152    &  \textbf{52.5 / 81.3}   &  -  &  57.2 / 85.6   &  47.7 / 71.1   &  \textbf{42.9 / 72.6 }  \\
resnet34    &  51.5 / 76.4   &  46.5 / 73.1   &  -   &  53.8 / 74.8   &  49.1 / 74.5   \\ \hline
vgg13\_bn  &  24.1 / 49.2   &  14.3 / 33.5   &  \textbf{25.1 / 54.1}   &  -  &  90.6 / 96.4   \\
vgg19\_bn    &  \textbf{27.1 / 57.5}   &  16.7 / 41.6   &  \textbf{27.6 / 60.7}   &  92.0 / 96.1   &  -  \\ \hline
\end{tabular}
\label{table: non-target-igsm}
\end{table}

\paragraph*{Ensemble based approaches}
In this part, we apply the ensemble-based attack proposed by \cite{liu2016delving}, which is rather effective in generating strong transferable adversarial examples. We examine whether we can further improve its performance  by employing our variance-reduced gradient. 

Both targeted
\footnote{Compared to non-targeted attack, we find that 
a larger step size $\alpha$ is necessary for generating strong targeted adversarial examples.
Readers can refer to Appendix A for more detailed analysis on this issue, though we cannot fully understand it. Therefore a much larger step size than the non-target attacks is used in this experiment.}
and non-targeted attacks are considered. 
For non-targeted attacks, the Top-1 success rates are nearly saturated; for targeted attacks, generating an adversarial example predicted by target models as a  specific label is too hard, resulting in a very small success rate.  Therefore, we instead adopt the the Top-5 success rate as our criterion to better reflect the improvement of method. A variety of model ensembles are examined, and the results are summarized in Table~\ref{tab: ensemble} (The corresponding Top-1 success rate can be found in Appendix).

As we can see, it is clear that variance-reduced attacks outperform the corresponding normal ones by a remarkable large margin for both non-targeted and targeted attacks.  More importantly, the improvement never be harmed compared to single-model case, which implies that \textbf{variance-reduced gradient can be effectively combined with ensemble method without compromise}.

%----------------------------------------------------

\begin{table}[!thb]
\renewcommand{\arraystretch}{1.2}
\centering
\caption{Top-5 success rates (\%) of ensemble-based approaches. The row and column denote the source and target model, respectively. For each cell, the left
	is the rate of normal method, while the right is that the variance-reduced counterpart.
	The corresponding Top-1 success rates can be found in Appendix B (Table~\ref{table: ensemble-igsm-top1}
	and Table~\ref{table: target-top1}).
}
\subtable[\textbf{Non-targeted attacks}: IGSM ($T=200, \alpha=1$) versus vr-IGSM ($T=10,\alpha=5, m=20, \sigma=15$), distortion
	$\varepsilon=15$.]{
\label{table: ens-nontarget-top5}
\begin{tabular}{c   |  c     c     c     c }\hline
      Ensemble                & densenet121   & resnet152    & resnet50     & vgg13\_bn      \\ \hline\hline
  resnet18+resnet34+resnet101 &  43.0 / 75.5   & 54.5 / 81.6	 & 62.6 / 85.4	  & 42.0 / 74.2  \\ \hline
vgg11\_bn +densenet161	      &  40.5 / 73.5    & \textbf{18.5 / 56.4}    & 33.4 / 70.2    & 68.3 / 85.6 \\ \hline
resnet34+vgg16\_bn+alexnet 	  &  26.5 / 65.2    & \textbf{15.7 / 55.3}    & 33.8 / 72.8    & 77.8 / 89.9 \\ \hline
\end{tabular}
}
\subtable[\textbf{Targeted attacks}: IGSM $(T=20,\alpha=15)$ versus vr-IGSM $(T=20,\alpha=15,m=20,\sigma=15)$, distortion
		$\varepsilon=20$. ]{
\label{table: target-top5}
\begin{tabular}{c   |  c     c     c     c }\hline
             Ensemble                 & resnet152    & resnet50     & vgg13\_bn    & vgg16\_bn  \\ \hline \hline
             resnet101+densenet121        & 28.1 / 56.8  & 26.2 / 52.4  & 7.7 / 23.6  &  8.1 / 29.7 \\ \hline
 \makecell{resnet18+resnet34+resnet101+densenet121}   & 50.4 / 70.4  & 54.7 / 72.4  &  23.2 / 44.2  & 28.1 / 52.6 \\ \hline
 \makecell{vgg11\_bn+vgg13\_bn+resnet18\\+resnet34+densenet121}   & 24.3 / 55.8  & 36.9 / 65.9  & - & 62.2 / 83.5 \\ \hline
\end{tabular}	
}
\label{tab: ensemble}
\end{table}

\paragraph{Momentum-based approaches}
Momentum-based attacks are recently proposed by \cite{dongboosting}, which won the first place in 
\emph{NIPS 2017 Non-targeted Adversarial Attack and Targeted Adversarial Attack} competitions. Here we compare our variance-reduced attacks with the momentum-based attacks. In this experiment, three networks of different architectures are selected.  The momentum decay factor $\mu=1$ is chosen as suggested in (\cite{dongboosting}), and for variance-reduced gradient, we use $m=20,\sigma=15$. All attacks are iterated for $T=5$ with step size $\alpha=5$. Table \ref{tab: momentum} reports the Top-1 success rates of non-targeted attacks of three attacks, including momentum-based IGSM (m-IGSM), vr-IGSM and momentum-based vr-IGSM (m-vr-IGSM). 

As shown in the table, our method outperforms momentum-based method significantly for all the cases. Furthermore, by combining with the variance-reduced gradient, the effectiveness of momentum-based method is improved significantly. 
\begin{table}[!thb]
\renewcommand{\arraystretch}{1.3}
\centering
\caption{
	Top-1 success rates(\%) of non-targeted attacks. The row and column denote the source and target model, respectively. Each cell contains three rates corresponding m-IGSM, vr-IGSM and m-vr-IGSM, attacks respectively. 
}
\vspace*{0.5em}
\begin{tabular}{c   |  c   |  c  |   c   }\hline
                    & resnet18  & densenet121   & vgg13\_bn            \\ \hline\hline
 resnet18  &  -  & 65.6 / 73.1 / 86.5	 & 70.4 / 77.7 / 86.7	  \\ 
densenet121 & 72.7 / 84.5 / 91.1   & -    & 68.7 / 80.3 / 86.7   \\ 
vgg13\_bn	  &  43.1 / 58.6 / 74.3    & 28.4 / 44.7 / 60.9    & -   \\ \hline
\end{tabular}
\label{tab: momentum}
\end{table}

\subsection{Influence of hyper parameters} 
\begin{figure}[!htb]
\centering
\subfigure[]{
	\raisebox{0mm}{\includegraphics[width=0.4\textwidth]{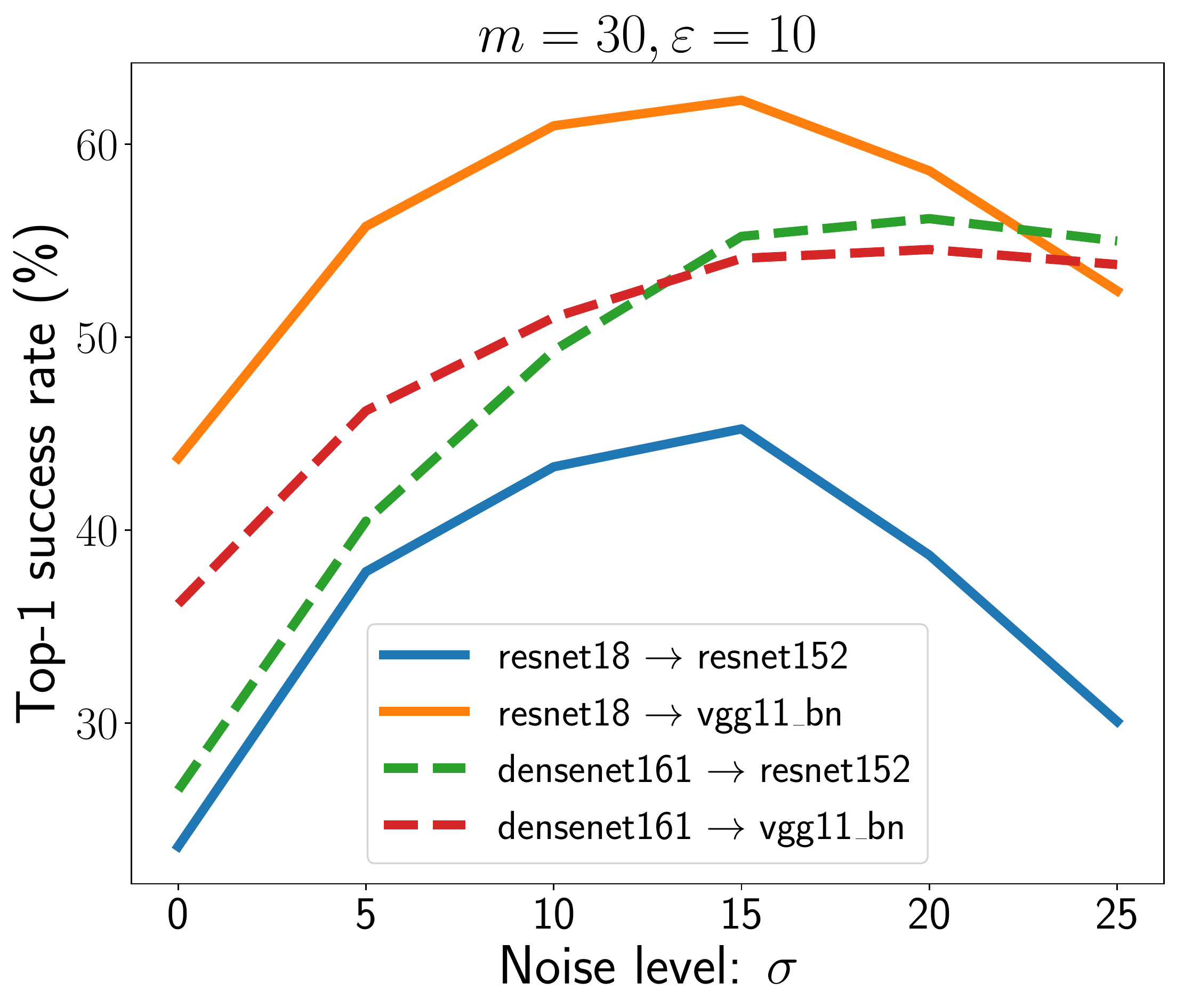}}
}
\hspace*{0.5em}
\subfigure[]{
	\includegraphics[width=0.45\textwidth]{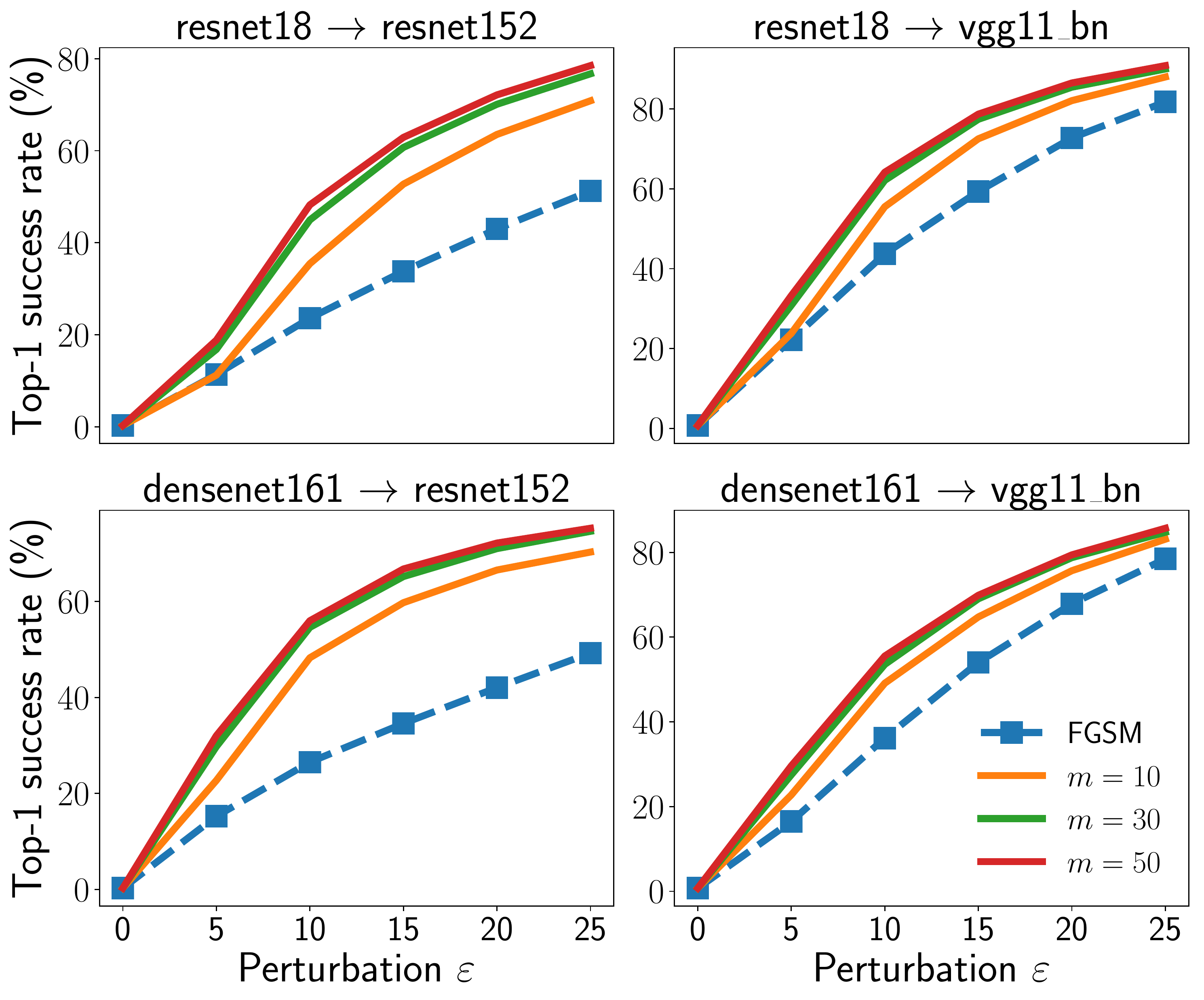}
}
\caption{
	\textbf{(a)} The sensitivity of the hyper parameter $\sigma$.
	\textbf{(b)} Success rates for nr-FGSM attacks with different $m$.
}
\label{fig: hyper-parameters}
\end{figure}

In this part, we explore the sensitivity  of hyper
parameters $m, \sigma$ when applying our variance-reduced gradient methods for
black-box attacks. We take ImageNet dataset as the testbed, and vr-FGSM attack
is examined. To increase the reliability,  four attacks are considered here.
The results are shown in Figure~\ref{fig: hyper-parameters}. It is not
surprising that larger $m$ leads to higher success rate for any distortion
level $\varepsilon$ due to the better estimation of the data-dependent
direction of the gradient. We find there is an optimal value of $\sigma$
inducing the best performance. Overly large $\sigma$ will introduce a large
bias in \eqref{eqn: vr-igsm}. Extremely small $\sigma$ is unable to smooth the landscape enough. 
Moreover the optimal $\sigma$ varies for different source models, and in this experiment it is about $15$ for \textit{resnet18},
compared to $20$ for \textit{densenet161}.

\subsection{Robustness analysis of adversarial examples}
Since variance-reduced attacks can be viewed as generating adversarial examples robust against Gaussian noise perturbation. Here we further examine their robustness to generic image transformations. The robustness is particularly important in practice, since it directly affects whether adversarial examples can survive in the physical world~(\cite{kurakin2016adversarial,athalye2017synthesizing,lu2017no}). To quantify the influence of transformations,
we use the notion of destruction rate defined by \cite{kurakin2016adversarial},
\[
	R = \frac{\sum_{i=1}^Nc(\bx_i)\left(1-c(\bx_i^{adv})\right) c(T(\bx_i^{adv}))}
			{\sum_{i=1}^N c(\bx_i)\left( 1 - c(\bx_i^{adv})\right)},
\]
where $N$ is the number of images used to estimate the destruction rate, $T(\cdot)$ is
an arbitrary image transformation. The function $c(\bx)$ indicates whether $\bx$ is
classified correctly:
\[
	c(\bx):=\left\{\begin{array}{rl}
		1,& \quad \text{if image $\bx$ is classified correctly} \\
		0,& \quad \text{otherwise}
	\end{array}\right.
\]
And thus, the above rate $R$ describes the fraction of adversarial images that are no longer misclassified after the transformation $T(\cdot)$.
\begin{figure}[!hb]
\centering
\includegraphics[width=1.0\textwidth]{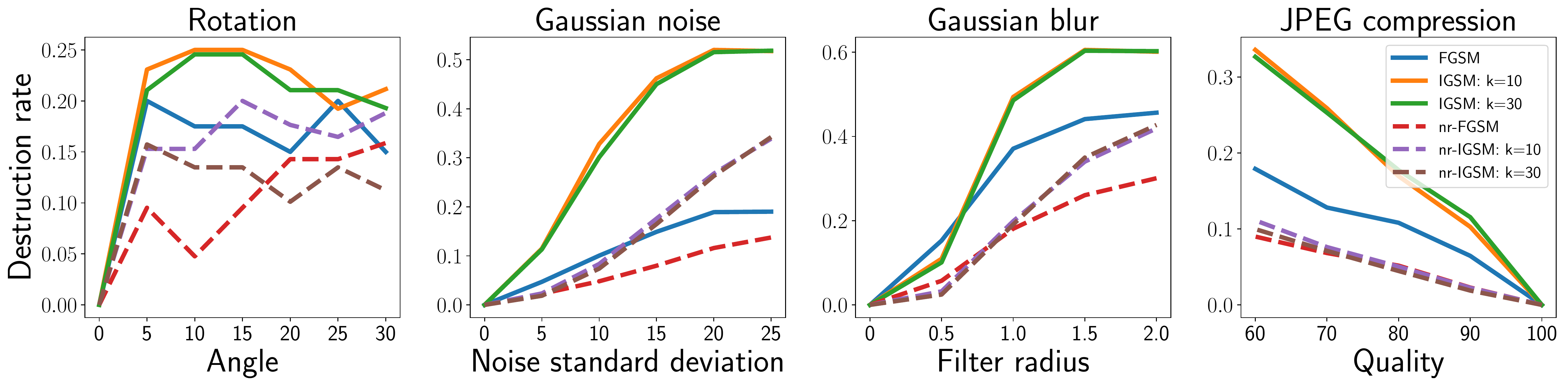}
\caption{Destruction rates of adversarial examples for various methods. For variance reduced attacks, we choose $m=20,\sigma=15$. The distortion $\varepsilon = 15$.}
\label{fig: robustness}
\end{figure}

\textit{Densenet121} and \textit{resnet34} are randomly chosen as our source  and target model, respectively; and four image transformations are considered: rotation, Gaussian noise, Gaussian blur and
JPEG compression. Figure~\ref{fig: robustness} displays the results, which show that adversarial examples generated by our methods are much more robust than those generated by vanilla methods. This numerical result is interesting, since we only explicitly increase the robustness against Gaussian noise in generating adversarial examples. This result suggests that the robustness of adversarial examples can also transfer among different image transforms.

%---------------------------------------------
\section{Discussion and Conclusion}
In this paper, we first investigated the influence of model-specific factors on the adversarial transferability. It is found that model architecture similarity plays a crucial role. Moreover models with lower capacity
and higher test accuracy are endowed with stronger capability for transfer-based attacks. 
we second demonstrate that the shattered gradient hinders the transfer of adversarial examples. 
Motivated by the understanding, we 
proposed the variance-reduced attack which can enhance the transferability of generated adversarial examples dramatically. Furthermore, the variance-reduced gradient can be combined with both ensemble and momentum based attacks rather effectively.
Lastly, we show that, the adversarial examples generated by variance-reduced attacks are much more robust than normal methods.

Our results imply that the robust (at least against Gaussian noise) adversarial examples tend to have stronger cross-model generalization, i.e. transferability. The similar phenomenon is also studied in  learning theory (\cite{bousquet2002stability,novak2018sensitivity}), i.e. the stable algorithm and hypothesis always tend to generalize better as well. 
So it is worth to further investigate if improving the robustness against generic transforms (\cite{athalye2017synthesizing}) can enhance the transferability as well. 
%---------------------------------------------

\bibliography{adv_examples}
\bibliographystyle{iclr2018_conference}
\newpage

%---------------------------------------------
% Appendix
%---------------------------------------------

\section*{Appendix A: Influence of step size for targeted attacks}
When using IGSM to perform targeted black-box attacks, there are two hyper parameters
including number of iteration $k$ and step size $\alpha$. Here we explore their
influence to the quality of adversarial examples generated. The success rates
 are calculated on $1,000$ images randomly selected according to description
of Section~\ref{sec: evaluation-method}. \textit{resnet152} and \textit{vgg16\_bn} are
chosen as target models. The performance are evaluated by the average Top-5 success rate
over the three ensembles used in Table~\ref{table: target-top5}.

\begin{figure}[!ht]
\centering
\includegraphics[width=0.7\textwidth]{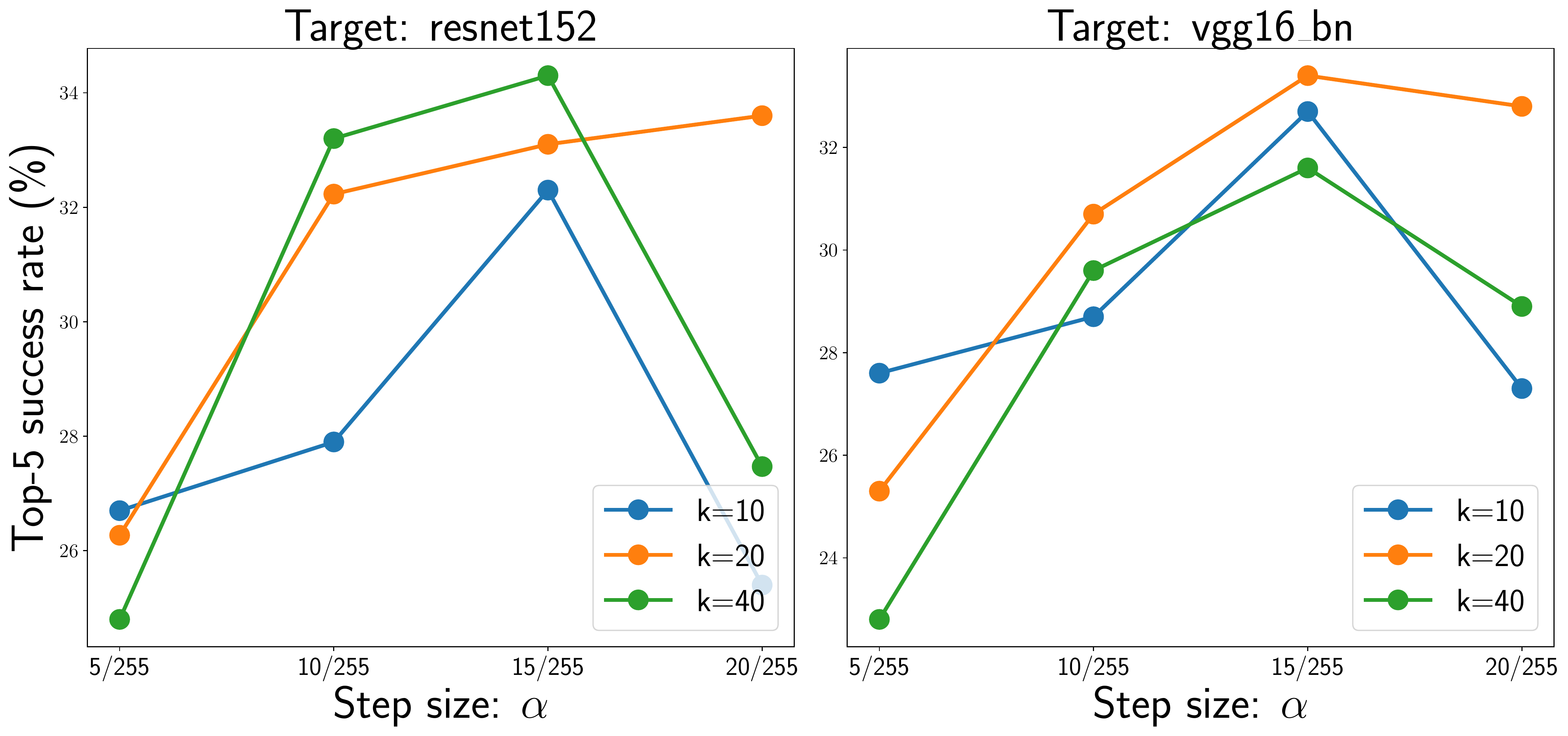}
\caption{Average success rates over three ensembles for different step size $\alpha$ and
number of iteration $k$. The three ensembles are the same with those in Table~\ref{table: target-top5}.
Distortion $\varepsilon=20$.}
\label{fig: stepsize-target}
\end{figure}

Figure~\ref{fig: stepsize-target} shows that for the optimal step size $\alpha$
is very large, for instance in this experiment it is about $15$ compared to
the allowed distortion $\varepsilon=20$. Both too large and too small step size
will yield harm to the attack performances. It is worth to note that with small
step size $\alpha=5$, the large number of iteration provides worse performance than
small number of iteration. One possible explanation is that more iterations lead the optimizer to
converge to a more overfitted solution.
In contrast, a large step size can prevent it and encourage the optimizer to explore more model-independent
area, thus more iteration is better.

\section*{Appendix B: Some additional experimental results}

\begin{figure}[!ht]
\centering
\includegraphics[width=0.3\textwidth]{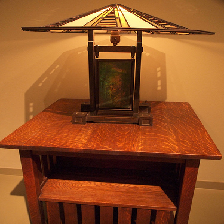}
\caption{The image used to perform decision boundary analysis. Its ID in ILSVRC2012 validation set is 26, with ground truth label being \textit{table lamp}.}
\label{fig: sample-image}
\end{figure}

\begin{table}[!ht]
\renewcommand{\arraystretch}{1.2}
\centering
\caption{
Top-1 success rates(\%) of non-targeted FGSM and vr-FGSM attacks between pairs of models. The row and column denote the source and target model, respectively. The left is the success rate of FGSM, while the right is that of vr-FGSM ($m=20,\sigma=15$).
 In this experiment, distortion $\varepsilon=15$.
}
\vspace*{0.5em}
\begin{tabular}{c   |   c     c     c   |  c     c}\hline
     		  & densenet121  & resnet152      & resnet34      & vgg13\_bn     & vgg19\_bn \\ \hline\hline
densenet121  &  - &    34.4 / 66.7   &  46.2 / 74.5 & 53.0 /  72.5  &  44.9 /71.4    \\
resnet152   & 39.2 / 67.2  &   -  &45.4 / 71.3     &  43.3 / 62.4   &  36.8 / 61.5     \\
resnet34   & 46.3 / 71.1   & 38.4 / 66.4 & -      &  54.4 / 70.5   &  46.7 / 68.8    \\\hline
vgg13\_bn  	 &23.0 / 48.4     &   16.0 / 34.4 &   28.0 / 53.2 &  -  &  54.2 / 84.6     \\
vgg19\_bn   & 28.1 / 58.8  & 18.7 / 46.4   & 31.5 / 62.3    &  62.2 / 87.1  &  -     \\ \hline
\end{tabular}
\label{table: non-target-fgsm}
\end{table}

\begin{table}[!ht]
\renewcommand{\arraystretch}{1.2}
\caption{
		Top-1 success rates (\%) of ensemble-based non-targeted IGSM and nr-IGSM attacks. The row and column denote the source and target model, respectively. The left is the success rate of
	IGSM ($T=100,\alpha=3$), while the right is that of vr-IGSM $(T=50,\alpha=3,m=20,\sigma=15)$. Since
	targeted attacks are much more difficult, we choose $\varepsilon=20$.
}
\centering
\vspace*{0.5em}
\begin{tabular}{c   |  c     c     c     c }\hline
      Ensemble                & densenet121   & resnet152    & resnet50     & vgg13\_bn      \\ \hline\hline
 resnet18+resnet34+resnet101  & 87.8 / \textbf{97.8}  & 94.6 / \textbf{98.9}  & 97.4 / \textbf{99.4}  &  84.1/96.1 \\ \hline
 vgg11\_bn+densenet161        & 86.8 / 97.2   & 62.9 / 89.7  & 80.3 / 94.8  & 94.9 / 98.4 \\ \hline
 resnet34+vgg16\_bn+alexnet   & 68.9 / 91.3   & 54.6 / 87.2 &  77.9 / 96.2  & 98.1 / \textbf{99.1}  \\ \hline
\end{tabular}
\label{table: ensemble-igsm-top1}
\end{table}

\begin{table}[!ht]
\renewcommand{\arraystretch}{1.2}
\centering
\caption{
	Top-1 success rates (\%) of ensemble-based targeted IGSM and vr-IGSM attacks. The row and column denote the source and target model, respectively. The left is the success rate of
	 IGSM ($T=20,\alpha=15$), while the right is that of vr-IGSM $(T=20,\alpha=15,m=20,\sigma=15)$. Since
	 targeted attacks are harder, we set $\varepsilon=20$.
}
\vspace*{0.5em}
\begin{tabular}{c   |  c     c     c     c  }\hline
       Ensemble                 & resnet152    & resnet50     & vgg13\_bn      & vgg16\_bn\\ \hline
 resnet101+densenet121         & 11.6 / 37.1   & 11.9 / 34.5  &  2.6 / 10.5    & 2.6 / 14.1      \\ \hline
 resnet18+resnet34+resnet101+densenet121    & 30.3 / \textbf{55.2}  & 36.8 / \textbf{57.3}  &  10.8 /\textbf{29.1}    & 12.8/35.0     \\ \hline
 \makecell{vgg11\_bn+vgg13\_bn+resnet18+\\
 resnet34+densenet121 }    & 10.1 / 35.1  & 22.2 / 47.9  & -   & 42.1/\textbf{72.1}    \\ \hline
\end{tabular}
\label{table: target-top1}
\end{table}

%====================================

\end{document}